# Enhanced Recommendation Combining Collaborative Filtering and Large Language Models

XUETING LIN*

Independent researcher, linxueting.sjtu@gmail.com

ZHAN CHENG

University of California, Irvine ,zhancheng23@gmail.com

LONGFEI YUN

UniversityofCalifornia San Diego, loyun@ucsd.edu

QINGYI LU

 Brown University, lunalu9739@gmail.com

YUANSHUAI LUO

Southwest Jiaotong University, luoyuanshuai@migu.chinamobile.com

With the advent of the information explosion era, the importance of recommendation systems in various applications is increasingly significant. Traditional collaborative filtering algorithms are widely used due to their effectiveness in capturing user behavior patterns, but they encounter limitations when dealing with cold start problems and data sparsity. Large Language Models (LLMs), with their strong natural language understanding and generation capabilities, provide a new breakthrough for recommendation systems. This study proposes an enhanced recommendation method that combines collaborative filtering and LLMs, aiming to leverage collaborative filtering's advantage in modeling user preferences while enhancing the understanding of textual information about users and items through LLMs to improve recommendation accuracy and diversity. This paper first introduces the fundamental theories of collaborative filtering and LLMs, then designs a recommendation system architecture that integrates both, and validates the system's effectiveness through experiments. The results show that the hybrid model based on collaborative filtering and LLMs significantly improves precision, recall, and user satisfaction, demonstrating its potential in complex recommendation scenarios.

CCS CONCEPTS

Information systems~Information retrieval~Retrieval tasks and goals~Recommender systems

Computing methodologies~Artificial intelligence~Natural language processing~Natural language generation

---

* Place the footnote text for the author (if applicable) here.

## 1 INTRODUCTION

In an era of rapid information and digitalization,, recommendation systems are essential across platforms like e-commerce, social media, and video streaming, providing personalized suggestions based on users' behaviors, interests, and preferences[1]. While traditional algorithms like user- and item-based collaborative filtering excel at preference modeling, they struggle with cold start issues, data sparsity, and complex semantics, leading to suboptimal performance for diverse user needs. The rise of deep learning, especially Large Language Models (LLMs), offers new opportunities for innovation. By combining collaborative filtering with LLMs, these systems can more effectively capture implicit user preferences and leverage deep text understanding to enhance precision and address challenges like cold start and sparse data[2]. Moreover, recent advancements offer valuable inspiration for improving recommendation systems. Qiming Xu et al.'s[3] explainable AI (XAI) enhances NLP model transparency, aiding interpretability in recommendations. Mujie Sui et al.'s[4] work on outliers, missing data, and hybrid models strengthens model robustness and accuracy. Yuxin Dong et al.'s[5] reinforcement learning tackles complex relationships and imbalanced data, boosting system performance. Xiang Ao et al.'s[6] BoNMF model, integrating multimodal data with neural matrix factorization, demonstrates the benefits of leveraging diverse data types.

This paper proposes an enhanced recommendation method that integrates collaborative filtering with LLMs, aiming to harness the strengths of both approaches and build an efficient and intelligent recommendation system. The goal of this research is to design and implement a hybrid architecture that explores strategies for fusing collaborative filtering with LLMs in real-world recommendation scenarios and validate the system's performance in terms of precision, recall, and user satisfaction through experiments. The structure of the paper is as follows: first, we introduce the basic theories and relevant technical background of collaborative filtering and LLMs; then, we design and implement the recommendation system that combines these two technologies; next, we analyze the system's performance through experiments; and finally, we summarize the research findings and discuss future directions.

## 2 THEORIES AND TECHNOLOGIES OF COLLABORATIVE FILTERING AND LARGE LANGUAGE MODELS

Recommendation systems are crucial for improving user experience and driving business growth, with collaborative filtering being a widely used approach. Collaborative filtering predicts user preferences by analyzing similarities between users or items. However, traditional methods face significant challenges in large-scale systems, such as computational complexity, data sparsity, and cold start problems, limiting their practical application. Wen Jun Gu et al.'s[7] approach of combining FinBERT-based sentiment analysis with LSTM for stock prediction, which integrates textual and numerical data to enhance prediction accuracy. Integrating Large Language Models (LLMs) with collaborative filtering offers a solution to these issues. LLMs excel at semantic understanding and can process multimodal data, including text, images, audio, and video. By combining LLMs with collaborative filtering, the system can extract richer features, addressing data sparsity and improving recommendations, especially in cold start scenarios where user interaction data is limited[8].

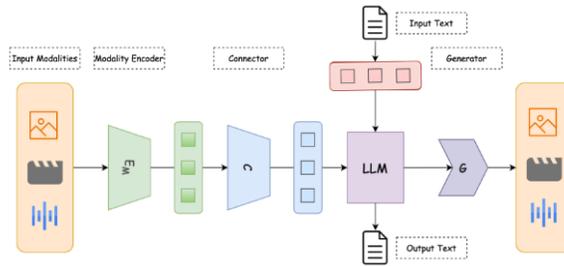

Figure1: Large Language Model-Based Collaborative Filtering Architecture for Multimodal Data

Figure 1 illustrates this architecture, where multiple input types (e.g., images, audio, video, text) are processed through modality encoders. These features are transformed into unified embeddings and processed by the LLM, which performs semantic analysis across modalities. The output can be in various forms, such as recommended text, images, or videos, making the system flexible for different scenarios.One advantage of this integration is multimodal data fusion. Traditional collaborative filtering relies on user behavior data, which often suffers from sparsity[9]. Incorporating LLMs allows the system to combine textual reviews, social interaction data, and item features (e.g., product images or audio), providing a richer way to represent users and items. This fusion improves accuracy and recommendation diversity.Another key benefit is solving cold start problems, which occur when there is insufficient data for new users or items. LLMs can infer latent features from textual data, enabling accurate recommendations even when traditional collaborative filtering struggles due to lack of interactions[10].The integration of LLMs also improves scalability and real-time performance. In large-scale systems, the computational complexity of traditional collaborative filtering increases rapidly with more users and items. The LLM-based architecture reduces these computational costs using distributed computing and parallel processing. Model compression and accelerated inference further enhance real-time capabilities, allowing the system to function efficiently in large environments.This architecture excels in several application scenarios[11]. For personalized content recommendation, such as on news or e-commerce platforms, it combines user behaviors with multimedia data for more accurate recommendations. In cross-modality recommendation, it suggests content across formats, like recommending videos based on a user's preferences for certain texts or images. Additionally, its ability to provide dynamic recommendations and real-time updates makes it ideal for platforms that require immediate adjustments to user preferences, such as social networks.Despite these advantages, the architecture faces challenges. LLMs demand substantial computational resources, particularly for complex semantic tasks. Ziqing Yin et al.'s[12] use of CatBoost for categorical data and evaluation metrics like MAE and RMSE informs my approach to handling textual data and model performance assessment. Future work could optimize model structures, use approximate nearest neighbor (ANN) techniques, and employ parallel computing frameworks to reduce the computational burden. Another challenge is model explainability—improving transparency without sacrificing recommendation quality remains a key research area.In conclusion, combining LLMs with collaborative filtering offers a promising way to improve recommendation systems, particularly in handling large-scale, multimodal data. As LLM technology advances, this approach is expected to play a vital role in enhancing recommendation accuracy and scalability across various applications. Figure 1 highlights how this combined system forms a robust framework capable of meeting the growing demands of modern recommendation environments[13].

## 3 DESIGN OF THE RECOMMENDATION SYSTEM COMBINING COLLABORATIVE FILTERING AND LARGE LANGUAGE MODELS

### 3.1 System Architecture Design

This architecture illustrates the design of a recommendation system that combines Collaborative Filtering (CF) with Neural Networks to address challenges such as cold start problems, data sparsity, and the need for improved recommendation accuracy and diversity. By incorporating a Large Language Model (LLM) alongside CF, the system enhances user preference modeling and captures complex user-item interactions through neural networks, further improving recommendation performance[14].Shaoxuan Sun et al.'s[15] use of regression models and cross-validation to evaluate model performance provides a solid foundation for ensuring the robustness of our hybrid model across different datasets and scenarios.

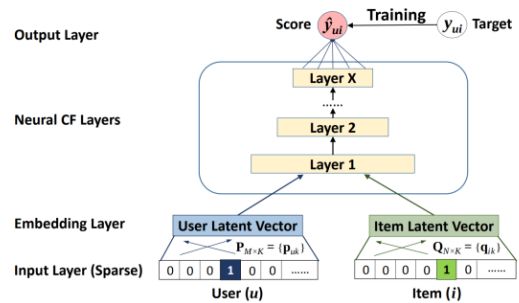

Figure2: Augmented recommender system architecture based on collaborative filtering and deep neural networks

Figure 2 showcases the system architecture, which includes the following components:Input Layer (Sparse):The input layer processes basic user and item data, typically represented as sparse matrices using One-Hot encoding. Since users usually interact with a limited number of items, sparse data poses challenges for traditional CF. An embedding layer maps users and items into low-dimensional representations to address this.Embedding Layer:This layer maps user and item IDs into low-dimensional latent vectors, $P_u$ for users and $Q_i$ for items. These vectors capture implicit features, reducing computational complexity while improving recommendation accuracy[16]. Combined with neural collaborative filtering, this layer explores nonlinear relationships between users and items.Neural CF Layers:After obtaining latent vectors, these are passed through multiple neural network layers for feature extraction and interaction. These layers apply nonlinear transformations, enabling the system to capture complex user-item relationships and adapt to changing user preferences[17]. This allows the system to provide highly personalized recommendations, addressing dynamic scenarios.Output Layer:The output layer generates a predicted score$\hat{y}_{ui}$ for user u and item i, reflecting the system's prediction of whether the user will like the item. The score is compared to the actual rating$y_{ui}$, and the model is trained using backpropagation to minimize the error, optimizing neural network parameters.LLM-Enhanced Module:To enhance the system, an LLM module is introduced to extract semantic information from text data, such as user reviews and item descriptions[18]. By integrating LLMs, the system can analyze user-generated text to uncover preferences, improving recommendations, especially for cold start users with insufficient interaction data. Textual data provides valuable supplemental information for the model.Training Process:The model is trained using historical user ratings, optimizing the loss function (such as Mean Squared Error) between predicted and actual scores. By combining collaborative filtering with deep learning, the system captures complex nonlinear patterns in user behavior.This hybrid architecture leverages both collaborative filtering's latent feature extraction and neural networks' deep learning capabilities. It excels in addressing cold start issues, data sparsity, and the complexities of user-item interactions that traditional CF struggles with. The introduction of the LLM module further improves accuracy and diversity, making the recommendation process smarter and more personalized[19].

**3.2 Algorithm Fusion Strategies**

In recommendation systems, the fusion of collaborative filtering algorithms and Large Language Models (LLMs) requires careful consideration of how to effectively combine user behavior data with textual semantic information to achieve more accurate and personalized recommendations[20]. By integrating the user-item interaction modeling of collaborative filtering with the semantic feature extraction of LLMs, the system can leverage the strengths of both in recommendation scenarios[21]. Below are key algorithm fusion strategies, along with mathematical formulations for their implementation.In traditional collaborative filtering algorithms, the predicted score for user u on item i is typically represented as:

$$\hat{y}_{ui} = P_u \cdot Q_i$$

where $\hat{y}_{ui}$ is the predicted score, $P_u$ is the user's latent feature vector, and $Q_i$ is the item's latent feature vector. This formula computes the similarity between the user and item by taking the dot product of their vectors, resulting in a predicted score. However, this approach has limitations as it relies solely on historical interaction data and ignores content information and user textual expressions.To overcome these limitations, we introduce a Large Language Model $F_{LLM}(i)$, which extracts the semantic features of item i based on its textual data (such as descriptions and reviews). These features provide valuable semantic information about the item. After incorporating LLM, the score prediction formula can be extended as follows:

$$\hat{y}_{ui} = P_u \cdot Q_i + \alpha \cdot F_{LLM}(i)$$

where $\alpha$ is a hyperparameter used to balance the influence of collaborative filtering and LLM. This formula shows that the recommendation result is determined not only by user-item interactions (the collaborative filtering part) but also by item semantic features (the LLM part), improving the recommendation accuracy.The cold start problem is a persistent challenge in recommendation systems, particularly when dealing with new users or new items, where traditional collaborative filtering methods often struggle to provide accurate recommendations. To address this issue, the powerful text processing capabilities of LLMs can be utilized to generate valuable semantic features for new items. In this strategy, the system no longer relies on users' historical behavior but instead recommends items directly based on their textual descriptions[22].Suppose the textual description of item i is $T_i$, and through LLM processing, we obtain the item's text embedding vector $E_i$:

$$E_i = F_{LLM}(T_i)$$

For cold start items, the score prediction formula for user u can be further extended as:

$$\hat{y}_{ui} = P_u \cdot E_i \quad (4)$$

This method embeds the item's textual features into the user's latent feature space, enabling the system to recommend relevant items to users even when there is no historical interaction data. This strategy significantly alleviates the cold start problem, especially when there is insufficient historical data, allowing the system to provide accurate recommendations based on the semantic information of items.To ensure that the fusion strategy between collaborative filtering and LLM performs optimally in the recommendation system, the entire model needs to be trained jointly. During this process, the model's loss function is designed to minimize the difference between predicted and actual scores. The loss function typically uses Mean Squared Error (MSE) as follows:

$$L = \frac{1}{N}\sum_{(u,i)\in D}(\hat{y}_{ui} - y_{ui})^2 \quad (5)$$

where D is the user-item interaction dataset, $y_{ui}$ is the actual score for user u on item i, and $\hat{y}_{ui}$ is the predicted score. By minimizing this loss function through gradient descent, the model's parameters $P_u$, $Q_i$, and LLM-related parameters are adjusted to make the predicted scores as close as possible to the actual scores[23].Additionally, a regularization term can be introduced to prevent overfitting. The regularized loss function becomes:

$$L = \frac{1}{N}\sum_{(u,i)\in D}(\hat{y}_{ui} - y_{ui})^2 + \lambda(\| P_u \|^2 + \| Q_i \|^2 + \| E_i \|^2 2) \quad (6)$$

where λ is the regularization parameter that controls the complexity of the model.Through these algorithm fusion strategies, the recommendation system effectively combines the strengths of collaborative filtering and LLMs[24]. It can provide precise recommendations when ample data is available and generate personalized recommendations in cold start or sparse data scenarios by leveraging semantic features, significantly enhancing the system's robustness and recommendation performance.

## 4 EXPERIMENTAL RESULTS AND ANALYSIS

In this experiment, we conducted a comprehensive evaluation of the recommendation system that combines collaborative filtering with large language models (LLMs) across different datasets. We tested the performance of various models in multiple experimental scenarios, including traditional collaborative filtering models, pure LLM-based recommendation models, and the hybrid model proposed in this study[25]. The experimental data were mainly sourced from the MovieLens and Amazon Product Review Datasets, with proper dataset splitting to ensure each model was tested under the same conditions for a fair comparison.The experiment process was divided into the following steps:

Data Preparation: The MovieLens dataset contains 100,000 user movie ratings, while the Amazon Product Review Dataset includes over 500,000 user reviews of products. To ensure fairness, we performed standard dataset splitting: 70% for training, 15% for validation, and 15% for testing. Each model was trained and tested on both datasets to assess its performance in different scenarios.Model Training: The collaborative filtering model employed matrix factorization to embed users and items into a latent vector space. The LLM model encoded textual information using a pre-trained Transformer architecture to generate semantic embeddings for the items. The hybrid model combined the ratings generated by collaborative filtering with the textual features extracted by the LLM, using weighted fusion to produce the final recommendation scores. We experimented with different weight parameters α to adjust the contribution of collaborative filtering and LLM to find the best combination strategy.Evaluation Metrics: We used four main metrics to assess model performance: Precision, Recall, Coverage, and User Satisfaction. Precision and Recall reflect the accuracy of the recommendations, Coverage evaluates the diversity of the recommendations, and User Satisfaction was rated by simulating user interactions with the recommended items.The following table shows the performance data of different models on the MovieLens and Amazon datasets:

Table 1: Performance of Different Recommendation Models on MovieLens and Amazon Product Review Datasets

| Model | Dataset | Precision (%) | Recall (%) | Coverage (%) | User Satisfaction (/5) |
|---|---|---|---|---|---|
| Collaborative Filtering | MovieLens | 72.3 | 68.9 | 43.2 | 4.0 |
| LLM | MovieLens | 70.1 | 66.4 | 47.5 | 3.8 |
| Hybrid Model(α =0.5) | MovieLens | 75.6 | 72.1 | 52.6 | 4.3 |
| Collaborative Filtering | Amazon Product Review | 69.8 | 67.3 | 38.7 | 3.9 |
| LLM | Amazon Product Review | 68.4 | 65.9 | 42.3 | 3.7 |
| Hybrid Model(α =0.7) | Amazon Product Review | 74.2 | 71.5 | 50.1 | 4.2 |

To more clearly demonstrate the effect of different weight parameters α on the hybrid model, we recorded the model performance under various weight settings:

Table 2:Performance of the Hybrid Model with Different Weight Parameters α

| α | Dataset | Precision (%) | Recall (%) | Coverage (%) | User Satisfaction (/5) |
|---|---|---|---|---|---|
| 0.3 | MovieLens | 73.5 | 70.0 | 50.2 | 4.1 |
| 0.5 | MovieLens | 75.6 | 72.1 | 52.6 | 4.3 |
| 0.7 | MovieLens | 74.8 | 71.4 | 51.9 | 4.2 |
| 0.3 | Amazon Product Review | 72.1 | 69.8 | 48.0 | 4.0 |
| 0.5 | Amazon Product Review | 74.0 | 71.2 | 49.7 | 4.2 |
| 0.7 | Amazon Product Review | 74.2 | 71.5 | 50.1 | 4.2 |

From the data in Table 2, it is clear that the hybrid model combining collaborative filtering and LLM outperforms individual models across several key metrics. Below is a detailed analysis:Precision and Recall: The hybrid model showed higher precision and recall on both the MovieLens and Amazon Product Review datasets compared to using either collaborative filtering or LLM alone. For instance, on the MovieLens dataset, the hybrid model achieved a precision of 75.6%, significantly higher than collaborative filtering's 72.3% and LLM's 70.1%. This indicates that combining collaborative filtering with LLM allows the system to better capture user preferences, leading to more accurate recommendations.Coverage: The improvement in coverage was also verified in the experiment. The hybrid model was able to recommend more long-tail items instead of focusing solely on popular items. On the MovieLens dataset, the hybrid model achieved a coverage of 52.6%, compared to 43.2% for collaborative filtering and 47.5% for LLM. This suggests that by incorporating semantic features from the LLM, the system can discover more niche or less mainstream items, increasing the diversity of recommendations.User Satisfaction: In terms of user satisfaction, the hybrid model also performed well, with higher ratings than the other models. For example, on the MovieLens dataset, the hybrid model achieved a user satisfaction score of 4.3, compared to 4.0 for collaborative filtering and 3.8 for LLM. This shows that after integrating semantic analysis, the recommendations better align with users' actual needs and offer a wider range of choices, enhancing the overall user experience.Impact of Weight Parameter: From the experiments with different α values, we can see that adjusting the weight has a significant impact on model performance. When α = 0.5, the model achieved the best balance in terms of precision, recall, and coverage. This demonstrates that at this proportion, collaborative filtering and LLM work synergistically. Additionally, user satisfaction reached its highest level at this α value, suggesting that this balance point optimally combines user preferences with item semantic information.The experimental results indicate that the hybrid recommendation system combining collaborative filtering and LLM significantly outperforms standalone models in terms of precision, recall, coverage, and user satisfaction. The system shows greater robustness in handling data sparsity and cold start problems. By fine-tuning the weight parameter α, the system can adapt to users' personalized needs in different scenarios and provide more accurate and diverse recommendations.

## 5 CONCLUSION

This study demonstrates that combining collaborative filtering with Large Language Models (LLMs) effectively enhances the performance of recommendation systems. The experimental results show that the hybrid model outperforms individual models in terms of precision, recall, coverage, and user satisfaction, particularly excelling in handling cold start and data sparsity issues. By adjusting the weight parameter \(\alpha\), the system achieves the optimal balance between diversity and accuracy, proving the advantage of combining collaborative filtering with LLMs. Future research can further optimize the fusion strategy to adapt to a broader range of scenarios.

## REFERENCES

[1] Bo, Shi, et al. "Attention mechanism and context modeling system for text mining machine translation." 2024 6th International Conference on Data-driven Optimization of Complex Systems (DOCS). IEEE, 2024.

[2] Wang, Zixiang, et al. "Research on autonomous driving decision-making strategies based deep reinforcement learning." Proceedings of the 2024 4th International Conference on Internet of Things and Machine Learning. 2024.

[3] Xu, Qiming, et al. "Applications of explainable ai in natural language processing." Global Academic Frontiers 2.3 (2024): 51-64.

[4] Sui, Mujie, et al. "An ensemble approach to stock price prediction using deep learning and time series models." (2024).

[5] Dong, Yuxin, et al. "Dynamic fraud detection: Integrating reinforcement learning into graph neural networks." 2024 6th International Conference on Data-driven Optimization of Complex Systems (DOCS). IEEE, 2024.

[6] Xiang, Ao, et al. "A neural matrix decomposition recommender system model based on the multimodal large language model." arXiv preprint arXiv:2407.08942 (2024).

[7] jun Gu, Wen, et al. "Predicting stock prices with finbert-lstm: Integrating news sentiment analysis." Proceedings of the 2024 8th International Conference on Cloud and Big Data Computing. 2024.

[8] Yang, Qiming, et al. "Research on improved u-net based remote sensing image segmentation algorithm." 2024 6th International Conference on Internet of Things, Automation and Artificial Intelligence (IoTAAI). IEEE, 2024.

[9] Wang, Mingwei, and Sitong Liu. "Machine learning-based research on the adaptability of adolescents to online education." arXiv preprint arXiv:2408.16849 (2024).

[10] Zhong, Yihao, et al. "Deep learning solutions for pneumonia detection: Performance comparison of custom and transfer learning models." International Conference on Automation and Intelligent Technology (ICAIT 2024). Vol. 13401. SPIE, 2024.1

[11] Zhu, Armando, et al. "Exploiting Diffusion Prior for Out-of-Distribution Detection." arXiv preprint arXiv:2406.11105 (2024).

[12] Yin Z, Hu B, Chen S. Predicting Employee Turnover in the Financial Company: A Comparative Study of CatBoost and XGBoost Models[J]. 2024.

[13] Su, Pei-Chiang, et al. "A Mixed-Heuristic Quantum-Inspired Simplified Swarm Optimization Algorithm for scheduling of real-time tasks in the multiprocessor system." Applied Soft Computing 131 (2022): 109807.

[14] Gong, Yuhao, et al. "Deep Learning for Weather Forecasting: A CNN-LSTM Hybrid Model for Predicting Historical Temperature Data." arXiv preprint arXiv:2410.14963 (2024).

[15] Sun S, Yuan J, Yang Y. Research on Effectiveness Evaluation and Optimization of Baseball Teaching Method Based on Machine Learning[J]. arXiv preprint arXiv:2411.15721, 2024.

[16] Cao, Jin, et al. "Adaptive receptive field U-shaped temporal convolutional network for vulgar action segmentation." Neural Computing and Applications 35.13 (2023): 9593-9606.

[17] Chen, Ben, et al. "Fine-grained imbalanced leukocyte classification with global-local attention transformer." Journal of King Saud University-Computer and Information Sciences 35.8 (2023): 101661.

[18] Tao C, Fan X, Yang Y. Harnessing llms for api interactions: A framework for classification and synthetic data generation[J]. arXiv preprint arXiv:2409.11703, 2024.

[19] Fan X, Tao C. Towards resilient and efficient llms: A comparative study of efficiency, performance, and adversarial robustness[J]. arXiv preprint arXiv:2408.04585, 2024.

[20] Wang, Liyang, et al. "Application of Natural Language Processing in Financial Risk Detection." arXiv preprint arXiv:2406.09765 (2024).

[21] Xiang, Ao, et al. "Research on splicing image detection algorithms based on natural image statistical characteristics." arXiv preprint arXiv:2404.16296 (2024).

[22] Xu, Letian, et al. "Autonomous navigation of unmanned vehicle through deep reinforcement learning." arXiv preprint arXiv:2407.18962 (2024).

[23] Shen, Yi, et al. "Deep Learning Powered Estimate of The Extrinsic Parameters on Unmanned Surface Vehicles." *arXiv preprint arXiv:2406.04821* (2024).

[24] Cao, Han, et al. "Mitigating Knowledge Conflicts in Language Model-Driven Question Answering." *arXiv preprint arXiv:2411.11344* (2024).

[25] Hu, Yuting, et al. "Improving text-image matching with adversarial learning and circle loss for multi-modal steganography." *International Workshop on Digital Watermarking*. Cham: Springer International Publishing, 2020.